# DEEP SEQUENCE MODELING:

# DEVELOPMENT AND APPLICATIONS IN ASSET PRICING

THIS VERSION: JUNE 10, 2020

**Lin William Cong, Ke Tang, Jingyuan Wang, and Yang Zhang**

**Lin William Cong** is the Rudd Family Professor of Management and Associate Professor of Finance at the Johnson Graduate School of Management at Cornell University in Ithaca, NY.
will.cong@cornell.edu

**Ke Tang** is a Professor of Economics in the School of Social Sciences at Tsinghua University in Beijing, China.
ketang@tsinghua.edu.cn

**Jingyuan Wang** is an Associate Professor in the School of Computer Science and Engineering at Beihang University in Beijing, China.
jywang@buaa.edu.cn

**Yang Zhang** is a Graduate Student in the School of Computer Science and Engineering at Beihang University in Beijing, China.
zhangyang96@buaa.edu.cn

**Abstract**

We predict asset returns and measure risk premiums    using a prominent technique from artificial intelligence:    deep sequence modeling. Because asset returns often exhibit sequential dependence that may not be effectively captured by conventional time-   series models, sequence modeling offers a promising path with its data-driven approach and superior performance. In this paper, we first overview the development of deep sequence models, introduce their applications in asset pricing, and discuss their advantages and limitations. We then perform a comparative analysis of these methods using data on U.S. equities. We demonstrate how sequence modeling benefits investors in general through incorporating complex historical path dependence    and that l   ong- and s   hort-term m   emory    based models tend to have the best out-of-sample performance.

THREE KEY TAKEAWAYS:

1- This paper provides a concise synopsis of deep sequence modeling with an emphasis on its historical development in the field of computer science and artificial intelligence. It serves as a reference source for social scientists who aim to utilize the tool to supplement conventional time- series and panel methods.

2- Deep sequence models can be adapted successfully for asset pricing, especially in predicting asset returns, which allow the model to be flexible to capture the high-dimensionality, nonlinear, interactive, low signal- to- noise, and dynamic nature of financial data. In particular, the model's ability to detect path- dependence patterns makes it versatile and effective, potentially outperforming existing models.

3- This paper provides a horse-race comparison of various deep sequence models for the tasks of forecasting returns and measuring risk premiums. Long short-term memory has the best performance in terms of out-of-sample predictive $R^2$, and long short-term memory with an attention mechanism has the best portfolio performance when excluding micro cap stocks.

# DEEP SEQUENCE MODELING:

# DEVELOPMENT AND APPLICATIONS IN ASSET PRICING

Deep learning, a powerful machine learning method, has seen great success in the field of artificial intelligence, primarily because of its ability to represent high-dimensional space and time dependence. However, standard feed-forward neural networks have limitations when observations are related in time or space, and may lose effectiveness with sequentially dependent input features (e.g. text, speech, asset prices). What we really need for sequence modeling is a neural network into which one can feed sequences of arbitrary length with elements indexed by time — that is, a memory system to remember events indexed by time steps in the past. Deep sequence modeling was proposed to address this limitation. It is related to the traditional time-series analysis in econometrics because both of them aim to uncover patterns in time-series data, but deep sequence models do so in a data-driven way, often entailing a large number of hidden states.

In finance, an emerging literature has applied machine learning in empirical asset pricing to predict asset returns or measure risk premiums (e.g., Feng, He and Polson 2018; Chen Pelgerz and Zhu 2020; Gu, Kelly, and Xiu 2020).[1] A typical approach relies on traditional feed-forward neural networks (Gu, Kelly, and Xiu 2020), taking a *one* step forecast with features of fixed dimensions while ignoring longer-term sequential dependency in asset prices/returns. Yet, asset returns are known to follow both short-term and long-term serial dependence, as manifested in return momentum and reversal patterns (De Bondt and Thaler 1985; Poterba and Summers 1988; Jegadeesh and Titman 1993, 2001; Asness, Moskowitz and Pedersen 2013). Therefore, modeling the sequential dependence of asset returns is a natural and important step forward in asset pricing, especially when designing strategies based on sequential dependence. We add to the literature by systematically exploring the mechanism and performance of various deep sequence modeling techniques, using the estimation of the risk premiums of U.S. equities for a concrete illustration.

We first overview the developments in sequence modeling using neural networks; starting from basic recurrent neural networks (RNNs) that consider sequential dependence relative to plain-vanilla deep neural networks (DNNs), we discuss the vanishing/exploding gradients problem, before discussing RNN variants, including Long- and Short-Term Memory (LSTM); gated recurrent unit (GRU); bi-directional LSTM (Bi-LSTM), which integrates both future and past time steps; and RNNs with an attention (ATT) mechanism (mimicking the mechanics of human vision). We also introduce the state-of-the-art sequence model, Transformer, with attention mechanism alone and which captures global dependencies despite dispensing with recurrence and convolutions.

---

[1] we follow studies such as Gu, Kelly, and Xiu (2020) to use the terms "expected return" and "risk premium" interchangeably, although one may be interested in distinguishing among the various components of expected returns in the literature. Our concept of "risk premium" is rather conventional in that regard: conditional expected stock returns in excess of the risk-free rate.

We then present, in the second part of the paper, a comparative analysis of all the deep sequence models in measuring equity risk premiums (forecasting equity returns). We conduct a large-scale empirical analysis and show that many deep sequence models have the ability to forecast asset returns. We cover nearly 20,000 individual stocks over 46 years from 1970 to 2016. We consider 54 input features each lagged up to 12 months before the month when returns are predicted; these include price-based signals (e.g., monthly returns), investment-related characteristics (e.g., changes in inventory over total assets), portability-related characteristics (returns on operating assets), intangibles (e.g., operating accruals), value-related characteristics (e.g., the book-to-market ratio), and trading frictions (e.g. the average daily bid-ask spreads). We establish the following empirical facts about the deep sequence models for return prediction.

We demonstrate the significant gains achieved by applying deep sequence models in asset pricing. Three deep sequence models (LSTM, GRU, Bi-LSTM) perform significantly better than baseline DNN model in out-of-sample $R^2$. LSTM achieves the highest $R^2$ of 0.45% which is a 10% improvement over DNN models. Moreover, equal-weighted portfolios constructed from all the deep sequence models we examine almost all have out-of-sample Sharpe ratio greater than 2 with monthly rebalancing, thus outperforming most models from earlier anomaly studies and are comparable with the best performers from recent machine-learning-based approaches (e.g., Cong et al., 2020). The efficacy of deep sequence models appears robust. For example, even after excluding microcaps, LSTM with an attention mechanism still achieves a Sharpe ratio of 2.03.

**Development of Deep Sequence Models**

In this section, we introduce the general structure of deep sequence modeling and then discuss various important deep sequence models developed in computer science and artificial intelligence (AI) disciplines over the past decades. To connect to financial economists, we use equity pricing applications as illustrations.

**Setup**

We first write an asset's excess return (the stock return minus the risk-free rate), $r_{t+1}^{(i)}$, as a sum of its prediction and an error term:

$$r_{t+1}^{(i)} = E[\mathrm{F}_t] + e_{t+1}^{(i)}, \tag{1}$$

where stocks are indexed as $i = 1, 2, ...$ and time (months) by $1, 2, 3, ...$ We use $F_t$ to denote information (filtration) up to $t$ when the expectation is conducted on. We also denote

$$E[r_{t+1}^{(i)}|F_t] = F(X^{(i)}|F_t), \qquad (2)$$

where $F(X^{(i)}|F_t)$ is a function of stock features $X^{(i)}$ up to $t$. The goal of our paper is to isolate $E[r_{t+1}^{(i)}|F_t]$ as a function of past input features (information) that have the best predictability for the realized return $r_{t+1}^{(i)}$. This function is thus represented by the deep sequence models, and therefore depends on features and time of the i$^{th}$ stock. For parsimony, we do not allow this function to depend on features of other stocks at any time in this paper. We introduce a cross-asset attention network in the companion paper (Cong et al. 2020) to capture cross-asset dependence.

Deep sequence models take a sequence of features of different stocks (including past returns) as inputs, which, in this paper, are denoted as $X = \{x_1, ..., x_t, ..., x_T\}$, where $x_t$ represents a set of features observed at $t$ (from 1 to T). Note that in this paper, we use bold lowercase letters to denote a vector and bold upper case letters for matrices. For a given stock $i$ at time $T$, our target is to predict its returns $r_{T+1}^{(i)}$ with historical information ($X^{(i)}$) up to $T$. With this setup, our deep sequence model consists of two parts:

1. Sequence representation: The sequence representation, $z^{(i)} = F(X^{(i)})$, is a vector to convert the raw input sequence into a representation (function) for downstream tasks. For example, when predicting returns $r_{T+1}^{(i)}$ with historical information ($X^{(i)}$) up to $T$, the downstream task is, thus the prediction of returns, whereas the sequence representation is a vector consisting of all input features up to T. We will introduce different forms of deep sequence models in the following sections.
2. Return forecasting: given the historical sequence representation, $z^{(i)}$, returns of asset $i$ can be predicted using a linear transformation $\hat{r}_{T+1}^{(i)} = w^T z^{(i)} + b$, where $w$ is a vector transforming sequence representation $z^{(i)}$ into output space.[2]

The sequence model is routinely trained on the training set by minimizing the loss function via gradient descent methodology, which is a first-order iterative (gradient) algorithm for finding the (local) minimum of a function (and hence to find the optimal model parameters). In our context, the minimization is on the sum of squared errors between the predicted and true excess return of the subsequent month. The loss function $L$ is thus written as:

---

[2] The linear functional form is for simplicity. It can be easily extended to other (non-linear) functional forms.

$$L = \sum_{(i,t)\subseteq T_{train}} (r_{t+1}^{(i)} - \hat{r}_{t+1}^{(i)})^2, \qquad (3)$$

where $\hat{r}_{t+1}^{(i)}$ is the estimated value from the model, and $T_{train}$ denotes the dataset for the training sample.

One common approach in computer science is the back-propagation through time (BPTT), proposed by Werbos (1990), wherein parameters update continuously via a going-backward scheme. Specifically, each of the neural network's weights receives an update proportional to the gradient (partial derivative) of the error function with respect to the current weights in each iteration of training. Therefore, a good (nonzero) gradient is the prerequisite for effectively updating parameters. Consequently, preventing the gradients from vanishing to zero or exploding to infinity are key to training sequence models of this kind.

**Recurrent Neural Network**

Humans have memories and often process sequential information to make decisions. In contrast, traditional feed-forward neural networks cannot deal with sequential inputs. RNNs were first proposed to address this issue (Elman 1990). RNNs introduce a notion of time to the plain-vanilla feed-forward neural network models by augmenting them with edges that span adjacent time steps (Lipton, Berkowitz, and Elkan 2015). Such a model architecture enables learning features and long-term dependencies (memories) from sequential and time-series data (Salehinejad et al. 2017). Exhibit 1 sketches a simple RNN structure and its unfolded representation. Essentially, nodes (neurons) in an RNN are connected to each other from a directed graph with a sequence, which allows RNNs to have a temporal behavior. These nodes are either input nodes ($x_t$ in Exhibit 1), output nodes (or results nodes, $y_t$, in Exhibit 1), or hidden nodes ($h_t$ in Exhibit 1). The hidden nodes are latent variables that are not observed directly but are considered as the key variable linking the inputs and outputs. The hidden nodes are connected sequentially, memorizing the historical information; therefore, outputs (fitted returns in our setup) can be linked to past outputs and inputs.

As shown in Exhibit 1, at time $t$, the input vector of stock features arrives at the input nodes, $x_t$, one at a time; the $h_t$ and $y_t$ are calculated as a function of the weighted sum of connected nodes. Equations 4 and 5 are the mathematical presentation of this relationship. In particularly, nodes with recurrent edges receive inputs from the current data point $x_t$ and from its previous hidden state $h_{t-1}$. The hidden state $h_t$ represents the memory of the units, and $y_t$ is the output at step $t$. Note that this structure enables the RNNs to store, remember, and process complex signals from the past information.

$$h_t = \tanh\ (W^{(hx)}x_t + W^{(hh)}h_{t-1} + b_h)\ , \qquad (4)$$

$$y_t = f_o(W^{(yh)}h_t + b_y), \qquad (5)$$

where tanh represents a tanh activation function[3] with inputs of the current stock features and past hidden state $h_{t-1}$, and $f_o$ is a function transferring from $h_t$ to the output values.[4] $W^{(hx)}$ and $W^{(yx)}$ are the corresponding weight matrices and $b_h$, $b_y$ are the biases. To obtain the forecast values at $T$, we do not need to calculate the output at each step from 1 up to T. We only need to take the hidden state at $T$, $h_T$, to find the output $y_T$, which is the estimated return at time T, $\hat{r}_{T+1}^{(i)}$, via equation 5. Therefore, the sequence representation is $z = h_T$ because the downstream work to estimate $y_T$ only depends on $h_T$. In other words, as long as we know $h_T$, we can compute the output.

The dynamics of RNNs across time steps can be visualized by *unfolding* it, as Exhibit 1 displays: Instead of interpreting the network as cyclic, one can view it as a deep network with one layer per time step and shared weights across time steps.

Exhibit 1: Simple RNN structure (left side) and its unfolded representation (right side)

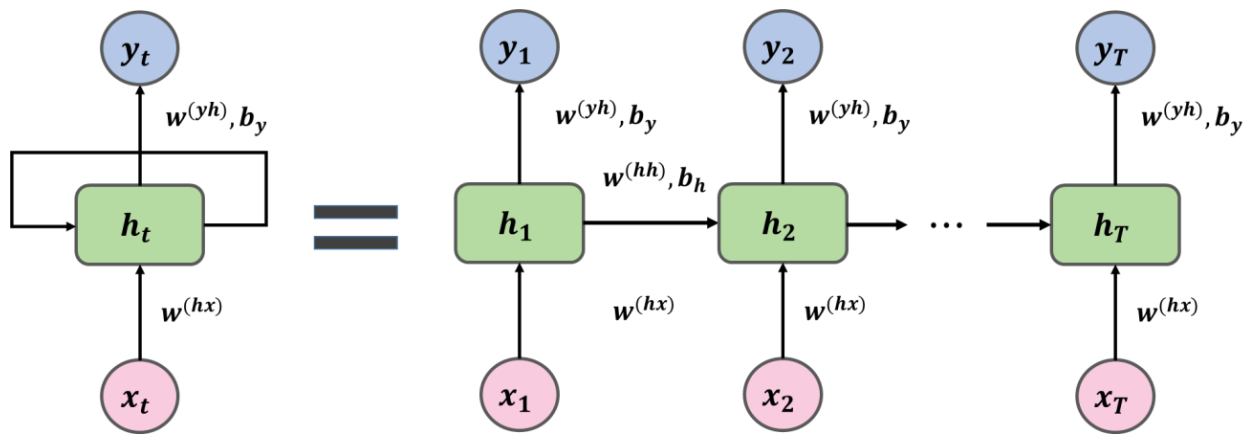

**Long Short-Term Memory RNN**

*Vanishing/Exploding Gradients:* Recurrent connections in RNNs can improve the performance of neural networks by leveraging their ability to understand sequential dependencies. However, the memory produced from recurrent connections may be severely limited by the algorithms employed for training RNNs. Because of the exponential growth and decay of the gradient, all models hitherto discussed fall victim to exploding or vanishing gradients during the training phase, resulting in the network failing to learn long-term sequential dependencies in data (Bengio, Simard, and Fransconi 1994). In machine learning, the vanishing gradient problem is a difficulty found in training artificial neural networks with gradient-based learning methods and back-propagation.

---

[3] A sigmoid function is a "S"-shaped, differentiable function, with resulting values going from 0 and 1. Many times, it determines the output of an algorithm to "yes" or "no." For example, the logistic function is a popular sigmoid function.
[4] Depending on the task, fo can take different forms; for example, fo is normally a softmax function for classification. It can also be an identity function for the purpose of predicting continuous returns.

In some cases, the gradient is too close to zero to dynamically change the weights. In the worst case, this may completely stop the neural network from updating and therefore cause it to fail training. Related is the exploding gradient problem, in which large gradients accumulate after iterating many steps; this tends to result in very large updates to weights, hence leading to an unstable network, which may make poor predictions or report nothing useful. Several models are specifically designed to efficiently tackle these problems; the most popular is the LSTM RNNs (Hochreiter and Schmidhube 1997).

Exhibit 2: Illustration of LSTM architecture

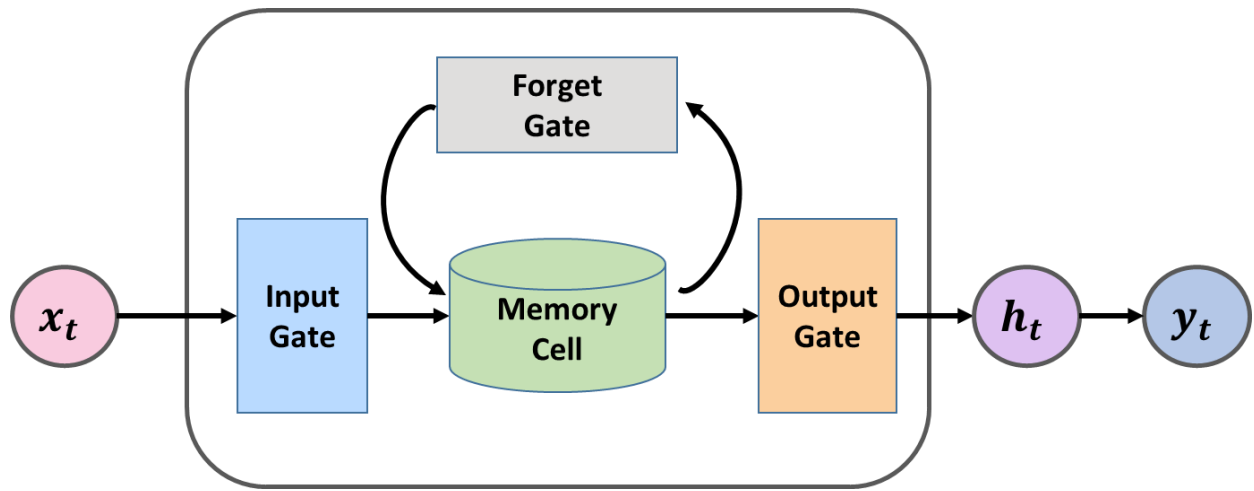

The LSTM adds a memory cell and associated functions (box in Exhibit 2) from $x_t$ to $h_t$ in an RNN and thus changes the structure of hidden units in RNNs from simple network layers to memory cells, wherein inputs and outputs are controlled by introducing gates (Exhibit 2). These gates control the flow of information to hidden neurons and preserve extracted features from previous time steps. The filtering procedure is to make a production of the gate value and the original information. For example, if the value of the gate is zero, information will not pass through; if it is one, the full information will go through. Specifically, the forget gate, as shown in Equation 6, is introduced to control the rate at which the value stored in the memory cell (value denoted as $c_t$ in Equation 9) decays. For periods when the input and output gates are off ($i_t = 0$ and $o_t = 0$) and the forget gate is not causing decay ($f_t$=1), a memory cell simply holds its value over time so that the gradient of the error stays constant during back-propagation over those periods (Lipton, Berkowitz, and Elkan 2015). This structure thus allows the network to potentially remember information for longer periods.

To be concrete, each unit in LSTM is computed as follows:

$$f_t = \sigma\big(W^{(fh)}h_{t-1} + W^{(fx)}x_t + b_f\big) \qquad (6)$$

$$i_t = \sigma(W^{(ih)}h_{t-1} + W^{(ix)}x_t + b_i) \tag{7}$$

$$o_t = \sigma(W^{(oh)}h_{t-1} + W^{(ox)}x_t + b_o) \tag{8}$$

$$c_t = f_t c_{t-1} + i_t \tanh\tanh(W^{(ch)}h_{t-1} + W^{(cx)}x_t + b_c) \tag{9}$$

$$h_t = o_t \tanh(c_t) \tag{10}$$

$$y_t = f_o(W^{(yh)}h_t + b_y), \tag{11}$$

where $\sigma$ is the sigmoid activation function, $tanh$ represents tanh activation function and $i, f, o$ are, respectively, values of the input, forget and output gates. Similar to the basic RNN, sequence representation for LSTM is $z = h_T$, as the downstream task (calculation of $y_T$) only depends on $h_T$.

**Gated Recurrent Unit**

**Although** LSTMs have been shown to be a viable option for avoiding vanishing or exploding gradients, they have a high memory requirement because the architecture contains multiple memory cells. Therefore, a simpler mechanism with a smaller number of parameters becomes attractive for less complicated tasks. GRU is such an algorithm. It adaptively captures dependencies of different time scales in GRUs. Similar to the LSTM unit (functions in the box of Exhibit 2), the GRU has gating units that modulate the flow of information inside the unit, but without having separate memory cells (Chung et al. 2014).

Specifically, a GRU contains an update gate and a reset gate. The update gate is responsible for determining which information from the previous memory to retain and for controlling the new memory to add. The reset gate allows the unit to forget the previous state. Compared with LSTM, GRU has a simpler architecture but fewer parameters. It is therefore designed for relatively simple tasks, and generally does not perform as well as LSTM for complicated tasks (Weiss, Goldberg, and Yahav 2018).

**Bi-D irectional LSTM**

Classical RNNs only consider past data in forward direction. Although simply looking at previous context is sufficient in many applications, it is also useful to explore the future context in speech recognition (Schuster and Paliwal 1997). Bi-LSTM is designed to deal with this problem.

A Bi-LSTM considers all available input sequences in both the past and future to estimate the output vector. Two RNNs are involved in the procedure: one processes the sequence from start to end in a forward time direction, and another processes the sequence backward from end to start in a negative time direction (Exhibit 3). Note that Bi-LSTM does not necessarily imply a look-ahead bias — as long as the information used is past information and forward-looking bias does not exist. For example, suppose the task is to forecast asset returns at $t = 10$; if the Bi-LSTM simply processes information from $t = 1$ to $t = 9$ in the first RNN and from $t = 9$ to $t = 1$ in the second RNN, it does not have a look-ahead bias. The Bi-LSTM's sequence representation is $z = concat(\vec{h}_T, \overleftarrow{h}_1)$ where $\vec{h}_T$ represents the forward vector with sequence in the forwarding direction and $\overleftarrow{h}_1$ represents the backward vector in the backward direction.

Exhibit 3: Bi-LSTM architecture

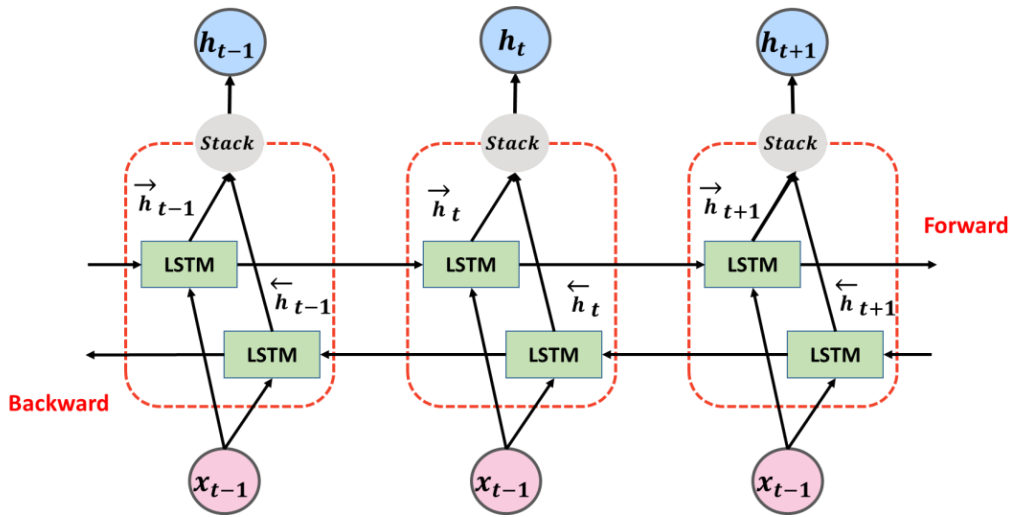

**RNNs with Attention**

In traditional RNNs, only the last hidden state $h_T$ (together with model parameters) is used to estimate the output $y_T$. Therefore, $h_T$ is regarded as containing all the abstract features in the entire sequence; any hidden states before $T$ are not involved directly in the calculation of $y_T$. However, such usage means the old information is washed out after being propagated over multiple time steps even when using LSTM (Hu, 2019). To address this issue, the attention mechanism is proposed to combine with the RNN (Chaudhari et al., 2019).

Overall, not only the hidden state at time T but also the hidden states before T all participate in the downstream tasks (i.e. forecasting the future returns in our paper, with various weights [scores]). Initially derived from human intuition and later adapted to machine translation for automatic token alignment, the attention mechanism, a simple framework that can be used for encoding sequence data based on the importance score each element is assigned, has become popular in computer science and AI. It has shown its significant improvements over prior approaches for a wide array of tasks (Bahdanau, Cho, and Bengio, 2014; Xu et al. 2015).

Specifically, to obtain a sequence representation $z$ which takes all the historical hidden states into consideration, one computes the attention scores with corresponding weights:

$$e_t = a(h_t, h_T) \tag{12}$$

$$\alpha_t = \frac{exp(e_t)}{\sum_i exp(e_i)} \tag{13}$$

$$z = \sum_t \alpha_t h_t, \tag{14}$$

where $\alpha_t$ is the weight of the hidden states, $a$ is the attention function that measures the similarity between step $t$ and the final step. One of the most commonly used attention function is called *additive attention* (Bahdanau, Cho, and Bengio 2014):

$$a(h_t, h_T) = v^T tanh(W^{(1)} h_t + W^{(2)} h_T), \tag{15}$$

where $v, W^{(1)}, W^{(2)}$ are parameters and $tanh$ is the activation function.

**Transformer**

Traditional RNN-based methods preclude parallelization within training examples, which becomes an issue with longer sequences. To this end, Google proposed a new network architecture, the Transformer, which relies solely on self-attention to compute representations of its inputs and outputs, dispensing with sequence-aligned RNNs or convolution entirely (Vaswani et al. 2017). The path lengths in RNNs and convolutional neural networks (CNNs) are linear and logarithmic, respectively. The path is constant in a Transformer, thereby making learning long-range dependencies easier. Moreover, the Transformer does not rely as heavily on the prohibitive sequential nature of input data as CNNs or RNNs do. The Transformer makes long-range dependencies in sequences easier to learn and allows for more parallelization.

The original Transformer follows an encoder-decoder architecture and achieves state-of-the-art performance in neural machine translation tasks. For asset pricing, we just use the Transformer encoder to encrypt the input sequence into its sequence representation. Cong et al. (2020) build on the Transformer encoder (Exhibit 4) and give a detailed description of the application of AI to portfolio management.

Exhibit 4: Illustration of Transformer Encoder

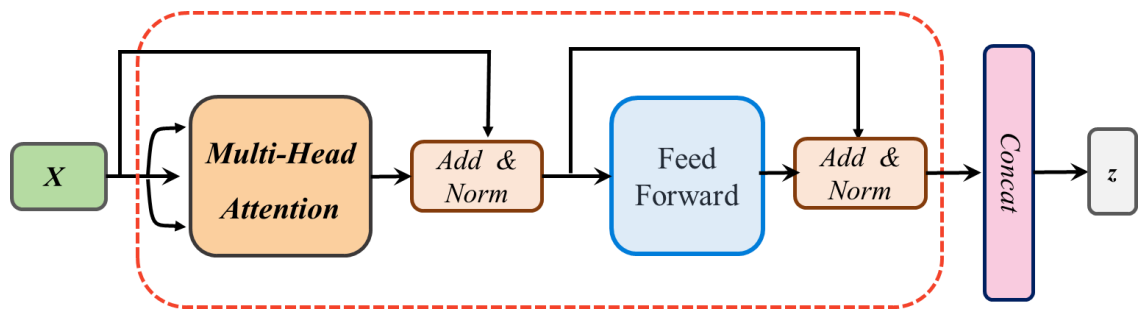

**Empirical Study using U.S. Equity Data**

The most important element in asset pricing is an understanding of the behavior of risk premiums of various assets; however, risk premiums are not directly observable and very difficult to correctly measure. Given that long-term serial dependences commonly exist in financial asset returns, this section applies different deep sequence modeling techniques to measure risk premiums of U.S. stocks. We also compare the deep sequence models with the traditional DNN models. Note that the DNN model belongs to the standard feed-forward neural networks.

*Data*: Monthly stock returns are from the Center for Research in Security Prices (CRSP). We follow the literature standard to focus on common stocks of firms in the United States and traded on Amex, Nasdaq, or NYSE. As for input features data, firms' balance-sheet data come from the S&P Compustat database. To mitigate survivorship bias due to backfilling, we also require that a firm appear in the dataset for at least one year.

Similar to Freyberger, Neuhierl, and Weber (2019), we classify the 51 firm characteristics into six categories: price-based signals (e.g., monthly returns), investment-related characteristics (e.g., changes in inventory over total assets), portability-related characteristics (returns on operating assets), intangibles (e.g., operating accruals), value-related characteristics (e.g., the book-to-market ratio), and trading frictions (e.g., the average daily bid-ask spreads). We consider lagged features up to 12 months before the month when returns are predicted. Overall, we have $12 \times 51$ input features to predict stock returns at any time. The features are listed in the Appendix.

Our sample period is from January 1970 to December 2016 with about 1.8 million monthly observations. The average number of stocks per month exceeds 3600. We divide the 47 years of data into 17 years of training samples (1970-1986) and the remaining 30 years of test samples (1987-2016). We perform rolling updates to fine-tune our model once every year.

*Evaluation Metrics*: To assess predictive performance for individual asset return forecasts, we calculate both out-of-sample mean squared error (MSE) and $R^2$, which are formulated as:

$$MSE = \frac{1}{|T_{oos}|}\sum_{(i,t)\in T_{oos}} \left(r_t^{(i)} - \hat{r}_t^{(i)}\right)^2 \quad (16)$$

$$R_{oos}^2 = 1 - \frac{\sum_{(i,t)\in T_{oos}} \left(r_t^{(i)} - \hat{r}_t^{(i)}\right)^2}{\sum_{(i,t)\in T_{oos}} \left(r_t^{(i)}\right)^2} \quad (17)$$

where the MSE and $R_{oos}^2$ summarize the prediction errors for different stocks and over time for the out-of-sample prediction.

*Model Configuration*: For the DNN model, we employ a standard feed-forward neural network with three hidden layers. The units/dimension of each layer is set as 256, 64, 8. For RNN- based deep sequence models (including basic RNN, LSTM, GRU, Bi-LSTM and LSTM-ATT), the layer number is set to 2, which is a common setting for such models. The dimensions of hidden states and corresponding weight matrices are set as 32. For Transformer, we adopt one Transformer Encoder block. We also reduce the dimension of embedding from 512 to 256 and reduce the dimension of feed-forward from 2048 to 64 for computational efficiency.

*Optimization*: We use the adam optimizer with initial learning rate $\gamma = 0.001$ to optimize our model. To avoid overfitting, we apply dropout to the output of both deep sequence models and DNN during training; the dropout rate is 0.2. We also add l2 regularizations on weight matrices and bias; the coefficient is set as 0.0005.

Exhibit 5 shows that three of six deep sequence models outperform the DNN model. LSTM achieves the best performance among all deep sequence models and the DNN, in terms of out-of-sample $R^2$ and MSE. It is not surprising that the basic RNN performs even worse than the baseline DNN. The gradient vanishing problem makes it difficult for the RNN to learn long sequential dependency, and it may converge to suboptimal model parameters. We also find that the Transformer model performs the worst among all these models, which is likely due to overparameterization. The architecture of Transformer is too complicated to make the model converge under this relatively simple asset-pricing task (compared with neural machine translation, for which the Transformer is originally adopted).

| **Exhibit 5: Comparison of Out-of-Sample MSE and $R^2_{oos}$** | | | | | | | |
|---|---|---|---|---|---|---|---|
| | DNN | RNN | LSTM | GRU | Bi-LSTM | LSTM-ATT | Transformer |
| **MSE (%)** | 12.61 | 12.63 | 12.59 | 12.59 | 12.59 | 12.59 | 12.63 |
| $R^2_{oos}$ **(%)** | 0.29 | 0.13 | **0.45** | 0.42 | 0.44 | 0.28 | 0.12 |
| LSTM-ATT: LSTM with attention mechanism | | | | | | | |

*Portfolio Performance*: We also construct portfolio management strategies based on these models. Note that portfolio performance provides an additional evaluation of the model and its robustness. Following the common practice in the literature, we study equal-weighted long-short portfolios which long the top 10% of stocks with the most positive predicted returns and short the bottom 10% of stocks with the most negative predicted returns each month. Note that following Freyberger, Neuhierl, and Weber (2019), we calculate turnover using $Turnover_t = \frac{1}{4}\sum_i \left|w_{t-1}^{(i)}\left(1 + r_t^{(i)}\right) - w_t^{(i)}\right|$, where $w_t^{(i)}$ represents the portfolio weight for stock *i* at timestep *t*. The coefficient $\frac{1}{4}$ is to avoid double counting (a factor of 2) and to adjust for that the long-short strategies have $2 exposure (another factor of 2). Portfolios constructed from attention-based algorithms such as LSTM-ATT and Transformer have much smaller turnovers than traditional factor portfolios and are smaller than in recent studies such as that by Freyberger, Neuhierl, and Weber (2019) as well.

Exhibit 6 shows the performance of both baseline DNN models and all six deep sequence models. Out-of-sample portfolio performance aligns very closely with the results on return predictability reported earlier. For an equal-weight portfolio constructed using all stocks, GRU achieves the greatest Sharpe ratio, 2.14. Note that when we exclude microcap stocks that fall below the 20th percentile of size, the performance (especially annualized return) of almost all these models drops, which is consistent with the findings of Avramov, Cheng and Metzker (2020). Nevertheless, the out-of-sample annualized return and Sharpe ratio are still sizable and significant. Exhibit 7 shows value-weighted portfolio performance; LSTM achieves the greatest Sharpe ratio (1.44) using all stocks. It is notable that although it does not perform so well in predictive $R^2_{oos}$, LSTM with an attention mechanism achieves the best performance in Sharpe ratio for both equal- and value-weighted portfolios when excluding micro-cap stocks. This result indicates that such a model is robust under plausible economic restrictions.

| Exhibit 6: Performance of Equal-Weight Portfolios for Different Deep Sequence Models (Annualized) | | | | | | | |
|---|---|---|---|---|---|---|---|
| Panel A: All Stocks | | | | | | | |
|  | DNN | RNN | LSTM | GRU | Bi-LSTM | LSTM_ATT | Transformer |
| Return (%) | 24.94 | 29.93 | 31.88 | 30.31 | 31.01 | 30.25 | 32.42 |
| Std.Dev(%) | 12.86 | 14.97 | 15.97 | 14.16 | 14.97 | 15.40 | 16.76 |
| Sharpe | 1.94 | 2.00 | 2.00 | **2.14** | 2.07 | 1.96 | 1.93 |
| Skewness | 3.67 | 2.56 | 2.57 | 2.02 | 2.45 | 2.59 | 2.21 |
| Kurtosis | 24.50 | 12.49 | 12.15 | 7.63 | 12.04 | 13.92 | 9.13 |
| Turnover(%) | 56.90 | 48.50 | 37.93 | 28.08 | 50.45 | 15.48 | 16.70 |
| MDD(%) | 7.73 | 3.52 | 5.11 | 4.83 | 5.49 | 6.79 | 7.10 |
| Panel B: Exclude Micro-caps(20% NYSE) | | | | | | | |
| Return (%) | 17.48 | 20.44 | 18.20 | 20.39 | 14.99 | 23.52 | 24.83 |
| Std.Dev(%) | 9.65 | 12.01 | 13.05 | 10.92 | 11.89 | 11.61 | 12.54 |
| Sharpe | 1.81 | 1.70 | 1.39 | 1.87 | 1.26 | **2.03** | 1.98 |
| Skewness | 4.69 | 3.10 | 2.51 | 3.62 | 2.81 | 3.89 | 2.74 |
| Kurtosis | 39.27 | 17.48 | 17.90 | 24.81 | 18.51 | 25.36 | 14.42 |
| Turnover(%) | 61.95 | 57.38 | 43.30 | 31.50 | 58.75 | 17.63 | 17.95 |
| MDD(%) | 2.39 | 2.76 | 15.43 | 2.63 | 10.72 | 3.54 | 2.33 |

Although not reported here, one can further control for various risk factors and impose economic restrictions. Deep sequence models are reasonably robust. Cong et.al. (2020) describe in detail the procedures involved for these robustness tests and for economically interpreting the models.

**Exhibit 7: Performance of Value-Weight Portfolios for Different Deep Sequence Models (Annualized)**

| Panel A: All Stocks | | | | | | | |
|---|---|---|---|---|---|---|---|
| | DNN | RNN | LSTM | GRU | Bi-LSTM | LSTM_ATT | Transformer |
| Return (%) | 20.83 | 3.83 | 27.08 | 23.75 | 31.95 | 33.33 | 43.51 |
| Std.Dev(%) | 24.75 | 10.43 | 18.75 | 57.58 | 25.34 | 26.03 | 34.71 |
| Sharpe | 0.84 | 0.37 | **1.44** | 0.41 | 1.26 | 1.28 | 1.25 |
| Skewness | 9.21 | 0.36 | 6.18 | 17.23 | 4.65 | 6.56 | 10.48 |
| Kurtosis | 101.86 | 5.21 | 63.39 | 313.19 | 35.08 | 59.39 | 150.68 |
| Turnover(%) | 56.43 | 33.13 | 31.50 | 19.25 | 52.40 | 11.60 | 13.90 |
| MDD(%) | 7.05 | 21.22 | 2.71 | 12.02 | 5.24 | 3.63 | 3.42 |
| Panel B: Exclude Micro-caps(20% NYSE) | | | | | | | |
| Return (%) | 12.24 | 11.03 | 17.25 | 19.49 | 13.13 | 15.95 | 24.08 |
| Std.Dev(%) | 8.78 | 10.34 | 10.83 | 29.77 | 15.48 | 8.89 | 14.25 |
| Sharpe | 1.40 | 1.07 | 1.59 | 0.65 | 0.85 | **1.79** | 1.69 |
| Skewness | 1.77 | 1.62 | 1.64 | 15.59 | -3.36 | 3.31 | 2.74 |
| Kurtosis | 8.80 | 8.81 | 6.06 | 273.14 | 47.64 | 22.44 | 13.90 |
| Turnover(%) | 57.58 | 45.08 | 38.68 | 23.55 | 59.33 | 11.50 | 14.28 |
| MDD(%) | 5.20 | 8.98 | 4.01 | 5.11 | 13.09 | 4.23 | 4.92 |

**Conclusion**

In this study, we provide an overview of the development of deep sequence modeling in the AI field and illustrate how it can be applied to asset pricing. We also demonstrate the estimation of asset risk premiums with the deep sequence models, including RNN, LSTM, Bi-LSTM, GRU, LSTM-ATT, and Transformer. Unlike traditional deep neural networks, deep sequence models incorporate historical sequential dependence on asset returns or input features (e.g., book-to-markets or dividends ratio). As Cong et.al. (2020) point out, by effectively accommodating the high dimensionality, nonlinearity, low signal to noise, feature interaction, fast dynamics, and sequence dependence seen in many economic data, deep sequence models complement traditional time-series models in a way similar to how neural network models complement regressions. With more flexible and versatile sequence modeling that is data driven, risk premiums are measured with less approximation and estimation error, which significantly helps identify reliable economic mechanisms behind various asset-pricing regularities.

Using the empirical context of return prediction and risk premium measurement as a proving ground, we perform a comparative analysis of methods in the sequence learning repertoire. We first introduce different deep sequence models in detail. We then analyze the performance of these deep sequence models and compare them with the traditional DNN models. Overall, the deep sequence models outperform the DNN models, owing to the inclusion of sequential dependence on the returns and features. LSTM and LSTM-ATT perform the best, and Transformer seems the most robust when economic restrictions are imposed (also demonstrated in Cong et.al., 2020). Overall, we demonstrate that deep sequence models benefit investors through better estimations of risk premiums and predictions of asset returns. Applications to other areas in business economics constitute interesting future work. At the highest level, our findings also help justify the growing role of AI in social sciences and the burgeoning fintech industry.

**Acknowledgement**: We are grateful to Frank Fabozzi, Andrew Karolyi, Andreas Neuhierl, Marcos de Prado, and Lu Zhang for their comments. We also thank Zihan Zhang for excellent research assistance.

# Appendix

This section details the construction of the 51 variables we use as input features. We obtain the raw data from three WRDS databases: CRSP, CRSP Compustat Merged, and Financial Ratio Firm Level. Underlined characteristics can be obtained from the Financial Ratio Firm Level database.

***A2ME***: We define assets-to-market cap as total assets over market capitalization.

***OA***: We define operating accruals as change in non-cash working capital minus depreciation scaled by lagged total asset.

***AOA***: We define AOA as absolute value of operation accruals.

***AT***: Total assets.

***BEME***: Ratio of book value of equity to market equity.

***Beta_daily***: Sum of the regression coefficients of daily excess returns on the market excess return and one lag of the market excess return.

***C***: Ratio of cash and short-term investments to total assets.

***C2D***: Cash flow to price is the ratio of income and extraordinary items and depreciation and amortization to total liabilities.

***CTO***: We define capital turnover as the ratio of net sales to lagged total assets.

***Dept2P***: Debt to price is the ratio of long-term debt and debt in current liability to the market capitalization.

***$\Delta ceq$***: The percentage change in the book value of equity.

***$\Delta(\Delta GM - \Delta Sales)$***: The difference in the percentage change in gross margin and the percentage change in sales.

***$\Delta So$***: Log change in the split adjusted shares outstanding.

***$\Delta shrout$***: Percentage change in shares outstanding.

***$\Delta PI2A$***: The change in property, plants, and equipment over lagged total assets.

***E2P***: We define earnings to price as the ratio of income before extraordinary items to the market capitalization.

***EPS***: We define earnings per share as the ratio of income before extraordinary items to shares outstanding.

***Free CF***: Cash flow to book value of equity.

***Idol_vol***: Idiosyncratic volatility is the standard deviation of the residuals from a regression of excess returns on the Fama and French three-factor model.

***Investment***: We define investment as the percentage year-on-year growth rate in total assets.

***IPM***: Pretax profit margin.

***IVC***: We define IVC as change in inventories over the average total assets of $t$ and $t-1$.

***Lev***: Leverage is the ratio of long-term debt and debt in current liabilities to the sum of long-term debt, debt in current liabilities, and stockholder equity.

***LDP***: We define the dividend-price ratio as annual dividends over price.

***MC***: Size is the market capitalization.

***Turnover***: Turnover is volume over shares outstanding.

**NOA**: Net operating assets are operating assets minus operating liabilities scaled by lagged total assets.

***NOP***: Net payout ratio is common dividends plus purchase of common and preferred stock minus the sale of common and preferred stock over the market capitalization.

***O2P***: Payout ratio is common dividends plus purchase of common and preferred stock minus the change in value of the net number of preferred stocks outstanding over the market capitalization.

***OL***: Operating leverage is the sum of cost of goods sold and sales, general, and administrative expenses over total assets.

***PCM***: The price-to-cost margin is the difference between net sales and costs of goods sold divided by net sales.

***PM***: The profit margin (operating income/sales).

***Prof***: We define profitability as gross profitability divided by the book value of equity.

***Q***: Tobin's Q is total assets, the market value of equity minus cash and short-term investments, minus deferred taxes scaled by total assets.

***Ret***: Asset excess return in the month.

***Ret_max***: Maximum daily return in the month.

*RNA*: The return on net operating assets.

*ROA*: Return-on-assets.

*ROC*: Ratio of market value of equity plus long-term debt minus total assets to cash and short-term investments.

*ROE*: Return on equity.

*ROIC*: Return on invested capital.

*S2C*: Sales to cash is the ratio of net sales to cash and short-term investments.

*Sale_g*: Sales growth is the annual percentage growth rate in annual sales.

*SAT*: We define asset turnover as the ratio of sales to total assets.

*S2P*: Sale to price is the ratio of net sales to the market capitalization.

*SGA2S*: SG&A to sales is the ratio of sales, general, and administrative expenses to net sales.

*Spread*: The bid-ask spread is the average daily bid-ask spread in the month.

*Std_turnover*: Standard deviation of daily turnover in the month.

*Std_vol*: Standard deviation of daily trading volume in the month.

*Tan*: We follow Hahn and Lee (2009) and define dene tangibility as ($0.715 \times$ total receivables + $0.547 \times$ inventories + $0.535 \times$ property, plant, and equipment + cash and short-term investments) / total assets .

*Total_vol*: Standard deviation of daily return in the month.